%% file: root.tex
\documentclass[letterpaper, 10pt, conference]{ieeeconf}  

\IEEEoverridecommandlockouts                              

\usepackage{amsthm}
\usepackage{times}
\usepackage{multicol}
\usepackage[bookmarks=true]{hyperref}
\usepackage{hyperref}
\usepackage{amsmath, amssymb}
\usepackage{amsfonts}
\usepackage{graphicx}
\usepackage{siunitx}
\usepackage{standalone}
\usepackage{booktabs}
\usepackage[ruled,vlined,linesnumbered,noend]{algorithm2e}
\usepackage{mdframed}
\usepackage{fancyvrb,multirow}
\usepackage{dsfont,mathabx}
\usepackage{array, booktabs}
\usepackage{makecell}
\usepackage[para,online,flushleft]{threeparttable}
\usepackage{amsmath,amssymb} 
\usepackage{pifont}  
\usepackage[most]{tcolorbox}
\usepackage{tabularx}
\usepackage{bm}
\usepackage{tikz}
\usepackage{pgfplots, pgfplotstable}
\usepackage{textcomp}
\usepackage{url}
\usepackage{verbatim}
\usepackage{lipsum}  
\usepackage{adjustbox}
\usepackage{subcaption}
\theoremstyle{definition}

\newcolumntype{C}{>{\centering\arraybackslash}X}
\newcolumntype{D}{>{\centering}p{5.2em}}
\newcolumntype{E}{>{\centering}p{3em}}
\newcolumntype{M}{>{\centering\arraybackslash}p{0.08\textwidth}}

\title{\LARGE \bf
  {DEXTER-LLM: Dynamic and Explainable Coordination
    of Multi-Robot Systems
in Unknown Environments via \\ Large Language Models}
}

\author{Yuxiao Zhu$^{2, *}$, Junfeng Chen$^{1, *}$, Xintong Zhang$^2$,
  Meng Guo$^1$ and Zhongkui Li$^1$
  \thanks{The authors are with $^1$the College of Engineering,
    Peking University, Beijing 100871, China;
  and $^2$the Division of Natural and Applied Sciences,
  Duke Kunshan University, Suzhou 215316, China. $^*$Equal contributions.
This work was supported by the National Natural Science Foundation
of China (NSFC) under grants 62203017, U2241214, T2121002;
and by the Ministry of Education under grant 2021ZYA05004.
Corresponding author: {\tt\small meng.guo@pku.edu.cn}.}
}

\begin{document}
\maketitle
\thispagestyle{empty}
\pagestyle{empty}

\input{contents/abstract.tex}
\input{contents/introduction.tex}
\input{contents/problem.tex}
\input{contents/solution.tex}

\input{contents/experiment.tex}

\input{contents/conclusion.tex}

\bibliographystyle{IEEEtran}
\bibliography{contents/references}

\end{document}

%% file: contents/abstract.tex
\begin{abstract}
  Online coordination of multi-robot systems in open and unknown environments faces significant challenges,
  particularly when semantic features detected during operation dynamically trigger new tasks.
  Recent large language model (LLMs)-based approaches
  for scene reasoning and planning primarily focus on one-shot,
  end-to-end solutions in known environments,
  lacking both dynamic adaptation capabilities for online operation
  and explainability in the processes of planning.
  To address these issues,
  a novel framework (DEXTER-LLM) for dynamic task planning in unknown environments,
  integrates four modules:
  (i) a mission comprehension module that resolves partial ordering
  of tasks specified by natural languages
  or linear temporal logic formulas (LTL);
  (ii) an online subtask generator based on LLMs
  that improves the accuracy
  and explainability of task decomposition
  via multi-stage reasoning;
  (iii) an optimal subtask assigner and scheduler that allocates subtasks
  to robots via search-based optimization;
  and (iv) a dynamic adaptation and human-in-the-loop verification module that implements multi-rate,
  event-based updates for both subtasks and their assignments,
  to cope with new features and tasks detected online.
  The framework effectively combines LLMs' open-world reasoning capabilities
  with the optimality of model-based assignment methods,
  simultaneously addressing the critical issue
  of online adaptability and explainability.
  Experimental evaluations demonstrate exceptional performances,
  with~$100\%$ success rates across all scenarios,
  $160$ tasks and $480$ subtasks completed on average
  ($3$ times the baselines),
  $62\%$ less queries to LLMs during adaptation,
  and superior plan quality ($2$ times higher) for compound tasks.
  Project page at \url{https://tcxm.github.io/DEXTER-LLM/}.
\end{abstract}


%% file: contents/introduction.tex
\section{Introduction}\label{sec:intro}

Heterogeneous multi-robot systems have demonstrated significant potential
in executing complex missions in unstructured, safety-critical environments,
such as disaster response and search-and-rescue operations~\cite{khamis2015multi}.
However, manual coordination of such fleets is often labor-intensive and
demanding for human operators.
To address this challenge, various task planning algorithms have been proposed that
automate the decomposition of tasks into subtasks, which are subsequently assigned to robots.
These approaches include model-based optimization~\cite{khamis2015multi,
  messing2022grstaps, liu2024time, chen2023accelerated}
and learning-based prediction~\cite{kannan2024smart, liu2024coherent, mandi2024roco}.
While these methods typically rely on manual inputs or predefined rules for task decomposition,
they are generally suited to small, well-defined environments.
In contrast, in open environments where the type, number, and distribution of
objects and features are unknown, such approaches become impractical.
In these settings, the strategy for task decomposition is unclear and
difficult to enumerate exhaustively. For example, strategies for fire suppression
depend on the available resources (e.g., water or fire extinguisher), victim rescue
strategies vary based on factors such as whether victims are buried under debris,
and the unknown number of victims to be rescued further complicates planning.
Therefore, the ability to reason and plan in real time, based on newly discovered features
in dynamic and open environments, is crucial.
This includes generating new tasks, evaluating different strategies for task
decomposition, and computing the sequence of subtasks to assign to robots.
Achieving this with guarantees on performance and correctness remains an open problem.

\begin{figure}[!t]
  \centering
  \includegraphics[width=0.99\linewidth]{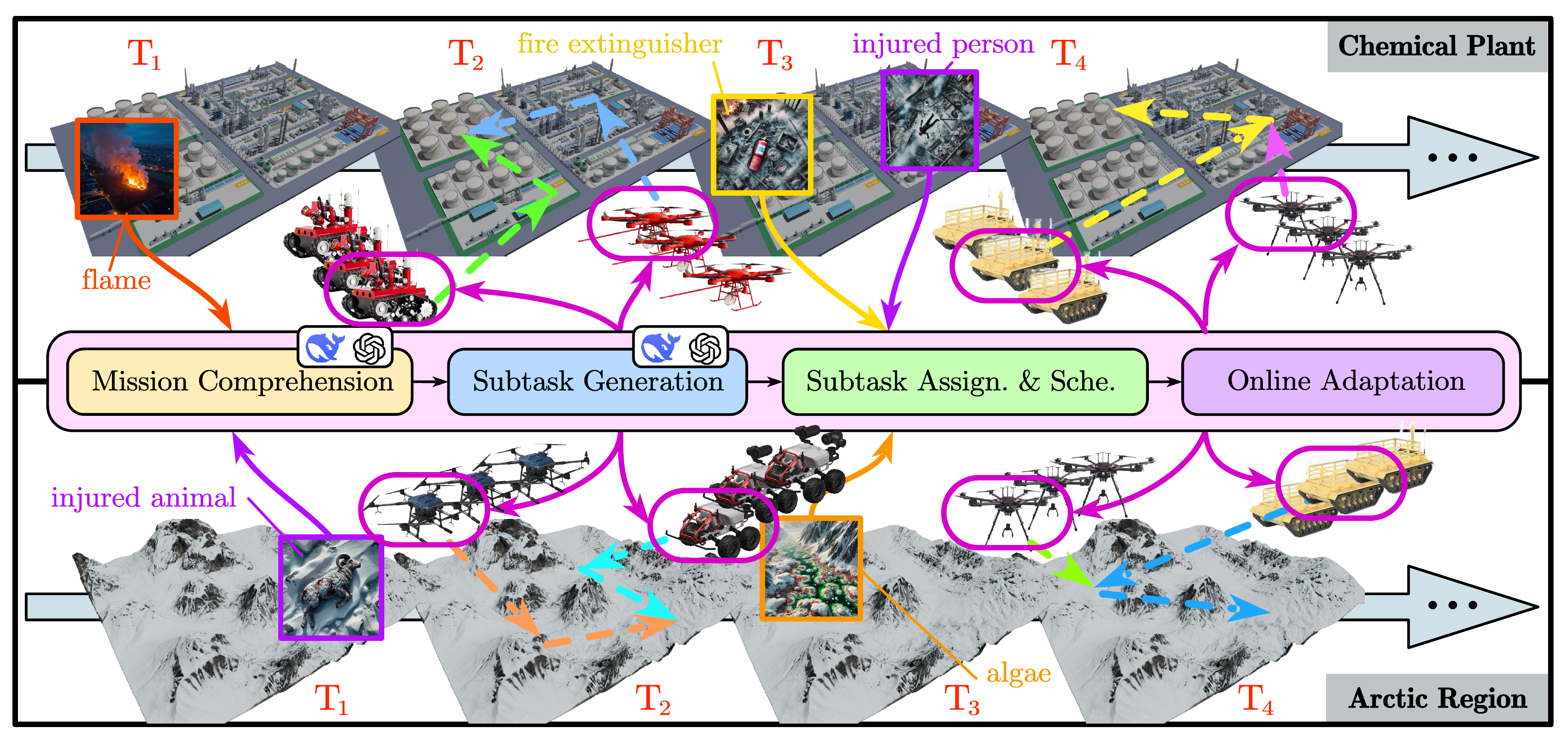}
  \vspace{-6mm}
  \captionsetup{font=footnotesize}
  \caption{\footnotesize Considered scenarios of fire suppression and wildlife protection,
    where the type and number of features (water, sand, algae...)
    and tasks (suppression, rescue...) are unknown;
    the strategies to decompose each task are reasoned
    (via multi-stage LLMs) and assigned (via optimization) online.
  }
  \label{fig:first}
  \vspace{-6mm}
\end{figure}
\subsection{Related Work}\label{subsec:rela-work}

\subsubsection{Task Planning in Robotics}
Long-term robotic tasks should be decomposed into a sequence of subtasks
that are executable by the deployed robots~\cite{khamis2015multi, messing2022grstaps}.
For static and known environments,
such decomposition can be achieved through classical methods,
such as: manual rules, PDDL formulations~\cite{aeronautiques1998pddl},
and linear temporal logic (LTL) specifications~\cite{guo2015multi}.
which require extensive domain expertise and exhibit limited adaptability
to open and dynamic environments due to hard-coded symbolic
representations~\cite{kloetzer2012ltl, guo2016task}.
Recent advances leverage LLMs to find potential strategies and moreover generalize
to different scenes.
Frameworks such as SayCan~\cite{ahn2022can} probabilistically map skills to subtasks,
while others translate natural language into structured models such
as PDDL~\cite{zhou2023iterative, capitanelli2024framework,zhou2024isr}
or and behavior trees~\cite{cai2025mrbtp}.
{
However, these methods mostly focus on single-robot scenarios,
often overlooking the complexities of multi-robot systems,
such as task allocation and coordination.
This limitation is particularly evident in dynamic environments
where the collaboration between robots is essential.
}
\subsubsection{LLM-based Multi-robot Task Planning}
The efficiency of multi-robot task planning relies on hierarchical architectures for mission prioritization,
task decomposition, and spatio-temporal scheduling~\cite{messing2022grstaps, luo2022temporal}.
While conventional optimization methods~\cite{khamis2015multi, chen2023accelerated} fail in open and dynamic environments due to rigid constraints~\cite{liu2024time},
LLM-based approaches~\cite{kawaharazuka2024real, chen2024autotamp} leverage commonsense reasoning
but suffer from insufficient formal verification~\cite{ahn2022can, gupta2025generalized, huang2022language, valmeekam2022large}
or guarantees on the quality of the generated plans.
Hybrid frameworks like ROCO~\cite{mandi2024roco} and SMART-LLM~\cite{kannan2024smart} integrate LLMs with optimization,
yet exhibit critical gaps: ROCO faces context explosion in large-scale deployments,
while SMART-LLM lacks adaptability to dynamic resource-task constraints.
There are three key limitations across existing systems,
e.g., LLM-GenPlan~\cite{silver2024generalized}, LiP-LLM~\cite{obata2024lip} and
aforementioned work:
open-loop architecture without verification or explainability,
inadequate response for various online events,
and missing failure recovery.
Although hierarchical methods like COHERENT~\cite{liu2024coherent} enable closed-loop recovery,
they cannot guarantee optimality for time-sensitive or long-term tasks.
Recent hybrid approaches combining LLMs with graph search, e.g., the generalized mission planner~\cite{gupta2025generalized},
improves adaptability but lacks real-time responsiveness and online recovery~\cite{ov2020impact}.
Therefore,
our proposed framework overcomes these limitations by integrating LLMs' open-world reasoning with model-based optimization methods,
enabling dynamic adaptation, human-in-the-loop verification,
and explainable planning for multi-robot coordination.
\begin{figure*}[!t]
  \centering
  \includegraphics[width=0.95\linewidth, height=0.3\linewidth]{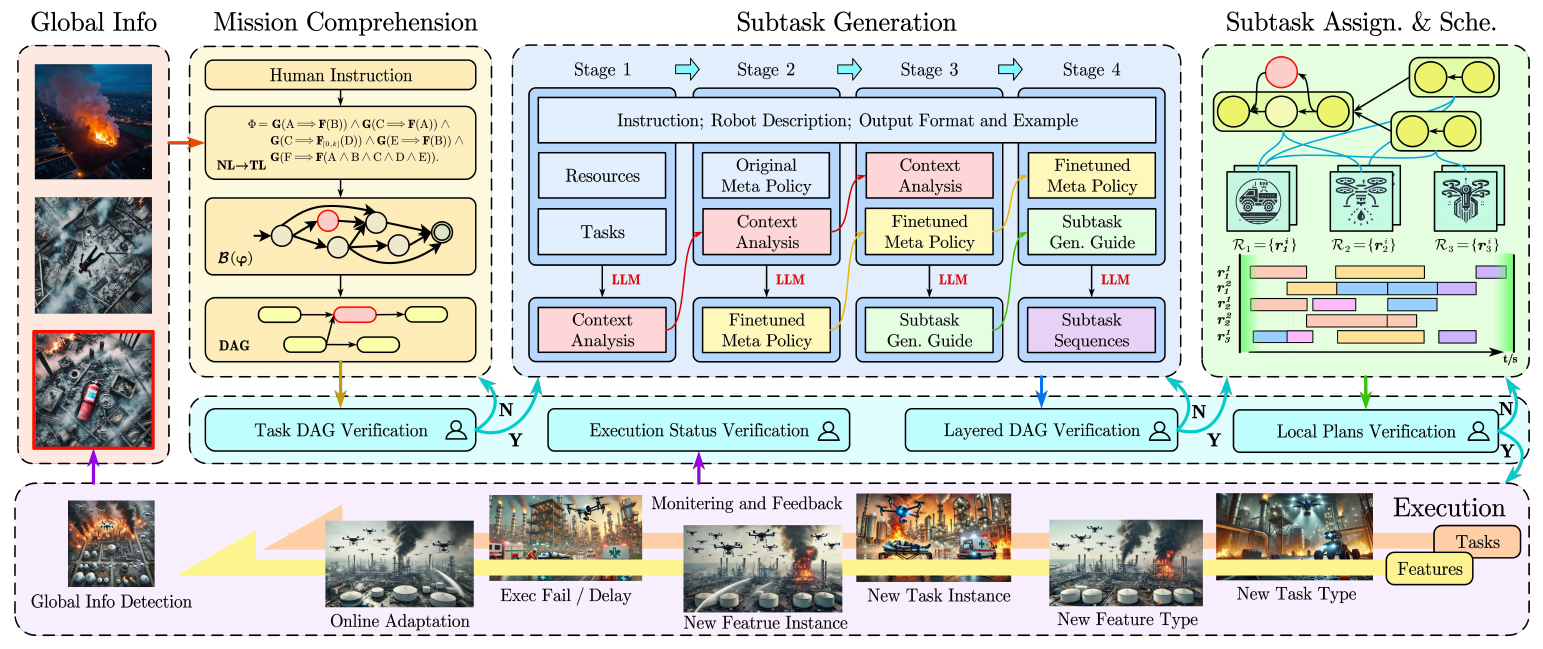}
  \vspace{-1mm}
  \captionsetup{font=footnotesize}
  \caption{\footnotesize Overview of the proposed framework,
    which includes four core modules:
    the abstraction of partially-ordered tasks
    from natural languages or LTL formulas (\textbf{left});
    the generation of plausible strategies to decompose each task
    into executable subtasks via LLMs (\textbf{middle});
    the subtask assignment and scheduling via model-based optimization (\textbf{right});
    and the human-in-the-loop online adaptation
    and verification (\textbf{bottom}).
  }
  \label{fig:framework}
  \vspace{-7mm}
\end{figure*}

\subsection{Our Method}\label{subsec:ours}
To overcome these limitations,
we propose~\textbf{DEXTER-LLM}—an acronym for \textbf{D}extrous \textbf{T}ask \textbf{EX}ecution and \textbf{R}easoning with Large Language Models—
an automated framework for dynamic task coordination of multi-robot fleets in open and unknown environments, as shown in Fig.~\ref{fig:first}.
It does not require any explicit modeling of potential semantic features
within the environment,
can handle newly-detected features and tasks online,
and can ensure correctness and optimality of the task assignment.
Specifically,
it first abstracts a set of partially-ordered tasks
from natural languages or LTL formulas.
Then, by incorporating the robot description and real-time scene,
LLMs generate the plausible strategies to decompose each task
into executable subtasks with temporal-logical constraints via multi-stage reasoning.
In addition, these subtasks encoded as layered directed acyclic graphs (DAGs)
are fed into a subtask scheduler that assigns sequences of subtasks
to each robot.
More importantly,
different from static and known environments,
a multi-rate adaptation scheme is proposed that reacts to
new features or tasks online by different modules at different events.
The proposed framework is shown to be generalizable owing to the
chained logic reasoning of LLMs,
verifiable by human feedback and the explicit constraints,
explainable via the multi-module structure and model-based optimization,
and scalable to large number of tasks.
Extensive numerical experiments are conducted to validate these properties
against different baselines in various scenarios.

Main contributions of this paper are two-fold:
(i) a versatile task coordination framework for heterogeneous multi-robot systems
in open and unknown environments,
which is generalizable, verifiable, explainable and scalable;
(ii) the seamless integration of LLMs and the model-based task planner,
which paves the way for new applications in open-world automated task planning.

%% file: contents/problem.tex
\section{Problem Description}\label{sec:problem}

Consider a fleet of heterogeneous robots that are capable of different
skills to interact with the environment, e.g., perception and manipulation.
The description of these skills are given as structured texts.
The environment is partially or fully unknown,
meaning that how the robots can interact with the environment is unknown.
Nonetheless, the fleet is given a high-level mission description
in natural language, which typically includes exploration and
other reactive tasks with different priorities
upon observing different features.

The planning objective is to: first comprehend the mission description
in terms of involved tasks and their temporal relations;
then generate and decompose the task into feasible subtasks,
according to the online observations;
assign the subtasks to robots to maximize task completion;
and react to execution delays, failures and new observations online.


%% file: contents/solution.tex
\section{Proposed Solution}\label{sec:solution}

\subsection{Overview of DEXTER-LLM}
\label{subsec:framework}

As shown in Fig.~\ref{fig:framework},
DEXTER-LLM is an online task planning framework enabling multi-robot systems
to dynamically generate subtasks through real-time environmental feedback.

\subsubsection{\textbf{Mission Comprehension}}

This module processes human-specified mission objectives,
such as: task types, priorities, temporal requirements,
by first translating them into verifiable LTL formulas~\cite{chen2023nl2tl}.
It then abstracts tasks through accepting paths in the nondeterministic B\"uchi automaton (NBA),
yielding a directed acyclic graph (DAG) where nodes denote tasks and edges enforce temporal constraints.

\subsubsection{\textbf{Subtask Generation}}

This module takes as inputs the set of tasks,
the global scene description, robot description, and historical logs.
Through multi-stage prompting, LLMs analyze scene semantics to derive: feasible task decomposition strategies;
and temporal-logic constraints, e.g., preceding, parallel.
The output is a \emph{layered} DAG where each task node expands into strategy-specific DAGs.

\subsubsection{\textbf{Subtask Assignment and Scheduling}}

This module processes the layered DAG,
environmental states (explored map),
and system status (robot conditions, task and feature distributions).
Employing branch-and-bound search with integer programming,
it optimizes subtask allocation and schedule
while ensuring the aforementioned constraints
and minimizing the overall makespan.
The output specifies the local plan of each robot as the timed sequence of
actions with precedence constraint at specific locations.

\subsubsection{\textbf{Online Adaptation and Human-in-the-loop Verification}}

This module integrates the aforementioned three modules
to handle real-time contingencies by processing dynamic inputs,
including new features, emergent tasks, execution status updates,
and robot failures.
Execution updates are verified by human operators and fed back to update local plans.
New types and instances of different features and tasks
are handled by different modules for efficiency.

\input{contents/solution/taskplan.tex}
\input{contents/solution/subtask.tex}
\input{contents/solution/allocator.tex}

\input{contents/solution/mechanism.tex}

%% file: contents/solution/taskplan.tex
\subsection{Module of Mission Comprehension}
\label{subsec:task-planning}

\subsubsection{Natural Language to LTL Formulas}
For intuitiveness, the team-wise mission description can be given
in natural languages, following certain templates.
Namely, it should contain how the system should behave in the nominal scenario,
and how the system should react upon new observations.
For explainability and non-ambiguity,
this description can be translated into LTL formulas as follows:
\begin{equation*}\label{eq:reactive-task}
\varphi \triangleq \varphi_\texttt{norm}
 \bigwedge_{\bar{e}\in \bar{E}} \Box \left( \varphi_\texttt{obs}^{\bar{e}}
  \rightarrow \Diamond\varphi_\texttt{rep}^{\bar{e}} \right),
\end{equation*}
where~\(\varphi_\texttt{norm}\) is the nominal task;
$\bar{E}$ is set of events that can be triggered by online observations;
$\varphi_\texttt{obs}^{\bar{e}}$ is a propositional formula over the same propositions
indicating event~$\bar{e}\in \bar{E}$;
and~$\varphi_\texttt{rep}^{\bar{e}}$ is the associated response.
Specifically, if the observed proposition~$\varphi_\texttt{obs}^{\bar{e}}$ is fulfilled,
the response task~$\varphi_\texttt{rep}^{\bar{e}}$ should be fulfilled.
The full semantics and syntax of LTL are omitted here for brevity; see e.g., \cite{baier2008principles}.
The \texttt{NL2TL} translator from~\cite{chen2023nl2tl} is adopted
to derive the mission formulas~$\varphi$, with some domain-specific modification.

\subsubsection{DAG of Tasks}
Once the mission formula~$\varphi$ is derived,
a relaxed partially ordered set (R-poset) is proposed
in the our earlier work~\cite{liu2024time},
to abstract the essential temporal constraints among the tasks.
In particular,
  the R-poset for~$\varphi$ is a
  3-tuple~$\mathcal{P}_\varphi=(\Omega_\varphi,\preceq_\varphi,\neq_\varphi)$:
	(I)~$\Omega_\varphi$ is the set of {tasks};
        (II)~$\preceq_{\varphi}\subseteq \Omega_{\varphi} \times \Omega_{\varphi}$
        is the ``precedence'' relation:
	If $(\omega_h,\omega_{\ell}) \in \preceq_{\varphi}$,
	then $\omega_{\ell}$ can only be {started} after $\omega_h$ is started.
	(III)~$\neq_{\varphi}\subseteq \Omega_{\varphi} \times \Omega_{\varphi}$ is
        the ``exclusion'' relation: If
	$(\omega_h,\omega_{\ell}) \in \neq_{\varphi}$,
        then tasks~$\omega_h,\omega_{\ell}$
	cannot be executed simultaneously.
The complete set of R-posets~\(\mathcal{P}_{\varphi}\) is as expressive
as the original formula and the associated NBA~\cite{liu2024time}.
The algorithmic details are omitted here for brevity.
Lastly, for convenience, the R-poset is encoded as a DAG,
where nodes are tasks,
and the labeled and directed edges represent the relation of
precedence and exclusion.
For instance, a mission is described as:
\emph{``Fully explore the area. After inspection, always extinguish detected fires,
rescue detected victims, prioritize large fires over small ones, and severe victims over mild ones.''}
and the associated LTL formula is given by:
$
\phi = (\Box \lozenge \texttt{exp}) \land \Box \big( \texttt{insp} \land (\texttt{fire}
\rightarrow \lozenge\texttt{ext}) \land (\texttt{hm} \rightarrow \lozenge\texttt{res})
\land ((\texttt{hm} \land \texttt{fire}) \rightarrow \lozenge (\texttt{res} \land
\lozenge \texttt{ext})) \big) \land \Box \big( (\texttt{lf} \land \texttt{sf})
\rightarrow \lozenge(\texttt{ext\_sf} \land \lozenge \texttt{ext\_lf}) \big) \land
\Box \big( (\texttt{shm} \land \texttt{mhm}) \rightarrow \lozenge (\texttt{res\_shm}
\land \lozenge \texttt{res\_mhm}) \big),
$
where the propositions are related to observations
or skills of the robots.
The associated DAG is shown in Fig.~\ref{fig:framework}.

%% file: contents/solution/subtask.tex
\subsection{Module of Subtask Generation}
\label{subsec:subtask-gen}

\subsubsection{Guided Multi-stage Reasoning via LLMs}
Given the set of tasks derived from the module of mission comprehension,
it remains to be determined how each task can be accomplished by the robotic
fleet, particularly given the semantic features within the environment.
To handle open and unknown environments,
the logic and reasoning capabilities of LLMs are utilized,
especially the ability to perform deductive reasoning over multiple steps.
However, different from the majority of work~\cite{huang2022language}
that relies on one-shot inference via LLMs,
this work proposes a sequential multi-stage procedure
to guide the reasoning of LLMs,
which is shown to improve significantly the context awareness,
accuracy and consistency of the resulting decomposition.
More specifically, as shown in Fig.~\ref{fig:framework},
it follows three sequential stages:
(I) \textbf{{Context Analysis}}.
The purpose of this stage is to filter out past and irrelevant observations
or events received from the global scene description.
(II) \textbf{{Meta Policy Tuning}}:
Updates the meta policy from the filtered observations.
For instance, upon finding new type of resources,
new strategies are generated.
The updated meta policy then guides LLMs to perform more accurate reasoning.
(III) \textbf{{Subtask Guide}}:
Given the complexities in subtask sequencing,
such as temporal relationships and conflicts of resources,
directly generating a standardized sequence is challenging for LLMs.
The stage of subtask guide builds on the above outputs,
allowing the LLM to focus on task complexities and improve consistency.
(IV) \textbf{Subtask Sequences}: The strategies as sequences of subtasks
are generated and encoded as layered DAG and JSON formats,
and fed into the subsequent module for allocation.



\begin{figure}[t!]
\begin{tcolorbox}[colframe=black, colback=gray!10, coltitle=black,
    sharp corners=southwest, boxrule=0.5mm, width=0.99\linewidth, coltext=black,
    boxsep=1pt, halign=justify]
\vspace{-0.08in}
\small
\textbf{Instruction}:
\begin{itemize}
    \item \texttt{Role}: You are an expert in emergency rescue.
    \item \texttt{Mission}: You should utilize the facilities and robots you have to efficiently extinguish the fire and rescue the injured person.
    \item \texttt{What to do}: Carefully read the following info to understand the situation and fulfill the mission.
\end{itemize}

\textbf{{Robots Team Description}}:
\begin{itemize}
\item \texttt{operating\_drone}: has flexible arm and hand,
  and skills: lift and hold, lay down and manipulate.
    \item \texttt{transporting\_cart}: is used to transport people or items,
           and skills: carry and move to destination.
         \item \texttt{fire\_extinguishing\_drone}: is used to extinguish fires,
           and its water tank is empty in the beginning,
           and skills: refill water tank, spray water.
         \item \texttt{fire\_extinguishing\_cart}: is used to extinguish fires,
           and its water tank is empty in the beginning,
           and skills: use fire extinguisher, refill water tank, spray water.
\end{itemize}
\textbf{Context Analysis Example}: The primary task is to extinguish the fire,
identified as flame, and there is an injured person requiring rescue.
The environment includes a detected water reservoir.

\textbf{Resources}: fire\_extinguisher

\textbf{{Task}}: [small\_fire, large\_fire, mild\_injury, severe\_injury]
\vspace{-0.08in}
\end{tcolorbox}
\vspace{-0.1in}
\captionsetup{font=footnotesize}
\caption{\footnotesize Outline of the prompt for the stage of ``Context Analysis''.}
\label{fig:context}
\vspace{-7mm}
\end{figure}


\subsubsection{Design of Prompt}
The prompt for each  stage is designed with
a consistent core framework including instruction and robots description,
while incorporating stage-specific elements.
For the stage of ``{context analysis}'',
the prompt includes instruction for specifying the role, mission and guide of LLMs,
followed by the current environmental context such as tasks and resources,
and lastly an example for structured outputs.
For the stage of ``{meta policy tuning}'',
the prompt augments the core structure with the results from context analysis,
along with the original meta policy,
the adaptation rules.
The stage of ``{subtask guide}''
integrates the updated meta policy and the results from context analysis
into the core structure,
with an intermediate representation for task inter-dependencies.
It outputs a hierarchical task description with precedence constraints.
Finally, for the stage of ``{subtask sequences}'',
the prompt incorporates the subtask guide and the meta policy,
generating the layered DAG for human verification and JSON
for the module of subtask assignment.
This multi-stage design of prompts enables progressive refinement of
LLM reasoning through contextual grounding and policy-aware adaptation,
for downstream execution.
Some excerpts are shown in Fig.~\ref{fig:context} and Fig. ~\ref{fig:guide}.

\begin{figure}[t!]
\begin{tcolorbox}[colframe=black, colback=gray!10, coltitle=black,
        sharp corners=southwest, boxrule=0.5mm, width=0.99\linewidth, coltext=black,
        boxsep=1pt, halign=justify]
\vspace{-0.06in}
    \small
    \textbf{Instruction}: ...

    \textbf{{Robots Team Description}}: ...

    \textbf{Original Meta Policy}: ...

    \textbf{Context Analysis}: \textcolor{black}{Output of context analysis}

    \textbf{Tuned Meta Policy Example}: If a fire is detected and water is available, then:
    Prioritize deploying drones and carts to refill water tanks first, then spray water on fire.
    \vspace{-0.06in}
    \end{tcolorbox}
\vspace{-4mm}
    \begin{tcolorbox}[colframe=black, colback=gray!10, coltitle=black,
        sharp corners=southwest, boxrule=0.5mm, width=0.99\linewidth, coltext=black,
        boxsep=1pt, halign=justify]
\vspace{-0.06in}
    \small
    \textbf{Instruction}: ...

    \textbf{{Robots Team Description}}: ...

    \textbf{Context Analysis}: \textcolor{black}{Output of context analysis}

    \textbf{Tuning Meta Policy}: \textcolor{black}{Output of meta policy tuning}

    \textbf{Subtask Guide Example}: Candidate Subtask Sequence 1:
    Extinguish fire first with a water-filled drone, then coordinate rescue:
    an operating drone lifts the injured while a transport cart moves them to rescue station.
    \vspace{-0.06in}
    \end{tcolorbox}
\vspace{-4mm}
    \begin{tcolorbox}[colframe=black, colback=gray!10, coltitle=black,
        sharp corners=southwest, boxrule=0.5mm, width=0.99\linewidth, coltext=black,
        boxsep=1pt, halign=justify]
      \vspace{-0.06in}
    \small
    \textbf{Instruction}: ...

    \textbf{Robots Team Description}: ...

    \textbf{Tuning Meta Policy}: \textcolor{black}{Output of meta policy tuning}

    \textbf{Subtask Guide}: \textcolor{black}{Output of subtask guide}

    \textbf{Subtask Sequence Example}: Candidate Subtask Sequence 1: {
        Sub-Task 2: {
            robot type: fire extinguishing drone,
            action: spray water to {flame},
            done by same robot as: Subtask 1,
            dependencies: [Subtask 1]
        }
      }
      \vspace{-0.6\in}
    \end{tcolorbox}
    \vspace{-0.1in}
    \captionsetup{font=footnotesize}
    \caption{\footnotesize Outline of the prompts for subsequent stages.}
    \label{fig:guide}
    \vspace{-6mm}
\end{figure}

%% file: contents/solution/allocator.tex
\subsection{Module of Subtask Assignment and Scheduling}
\label{subsec:subtask-sched}

\subsubsection{Multi-robot Subtask Assignment
  under Temporal Constraints}
Given the layered DAG including the ordered tasks and
the associated strategy of subtasks,
it remains to decide which robots in the fleet should perform
which subtasks at what time,
in order to minimize overall time to accomplish all tasks.
In contrast to existing methods~\cite{liu2024time},
this assignment and scheduling problem has two key characteristics:
(I) the strategy of each task, namely the subtasks,
should be determined along with the assignment;
(II) both the tasks and subtasks have strict
temporal constraints including precedence and exclusion
or different priorities.
Moreover, the mapping from subtasks to the required
primitive skills is abstracted from the
``Sequence of subtasks'' contained in the output
of the previous module.
Lastly, the time duration of ``navigation'' is computed
based on the explored map and the robot velocity,
while the duration of the other skills should be specified
in the robot description.

\subsubsection{Branch-and-bound Search with Integer Optimization}
Due to the combinatorial complexity of the above assignment problem,
a hybrid approach is proposed that combines the branch-and-bound (BnB) search
with mixed integer optimization to determine the optimal assignment and schedule.
Compared with the one-shot inference via LLMs,
this approach is verifiable, explainable and provides performance guarantee.
More specifically,
a search tree is constructed via iterative node selection and expansion.
Each node is a partial assignment of system,
i.e.,~$\boldsymbol{\nu}=(\nu_1,\nu_2,\cdots,\nu_N)$,
where~$\nu_n$ is local plan of robot~$n$ as the sequence
of grounded skills (including the starting time, ending time
and location of each skill).
The root node is an empty assignment.
\textbf{Selection}: the node with the minimum makespan is selected
within the set of existing nodes.
\textbf{Expansion}: once a node is selected, it is expanded by
selecting one strategy of any unassigned task
and assigning the contained subtasks to the robots.
To avoid producing infeasible assignments, the partial-ordering constraints
in~$\mathcal{P}_{\varphi}$ should be respected,
i.e., a task can not be assigned before its preceding task in~$\preceq_{\varphi}$.
Once the strategy is selected, the set of subtasks to be assigned
is determined along with their temporal-logic constraints.

Then, to determine the optimal assignment and the schedule
for this node, a mixed integer program is formulated as follows:
(I) the variables include positive reals
for the starting and ending time of each subtask,
and integer variables to indicate the index of each subtask
within the local plan of each robot;
(II) the objective is to minimize the maximum ending time of all subtasks;
and (III) the constraints include
the capability constraint,
the temporal-logic constraint among the subtasks (formulated via the big-M
inequality),
the environment constraint including navigation and subtask duration.
Lastly, the integer problem is solved via off-the-shelf solvers,
e.g.,~\texttt{Gurobi}~\cite{GurobiManual},
yielding the optimal makespan for this node,
the subtask assignment and the associated schedule.
This node is called feasible if integer problem is feasible with
a finite makespan.
Then, the procedure of selection and expansion is repeated,
until the planning time elapsed or all combinations of possible strategies
for the tasks are exhausted in the search tree.
The lower bound on the makespan of each node
is computed as the summation of the minimum duration of all remaining tasks,
while the upper bound is given by a one-step greedy assignment algorithm.
Algorithmic details are omitted here due to limited space.

%% file: contents/solution/mechanism.tex
\subsection{Module of Online Adaptation and Verification}
\label{subsec:mechanism}

\subsubsection{Multi-level Online Adaptation} \label{subsubsec:adaptation}
\begin{figure}[!t]
    \centering
    \includegraphics[width=0.99\linewidth]{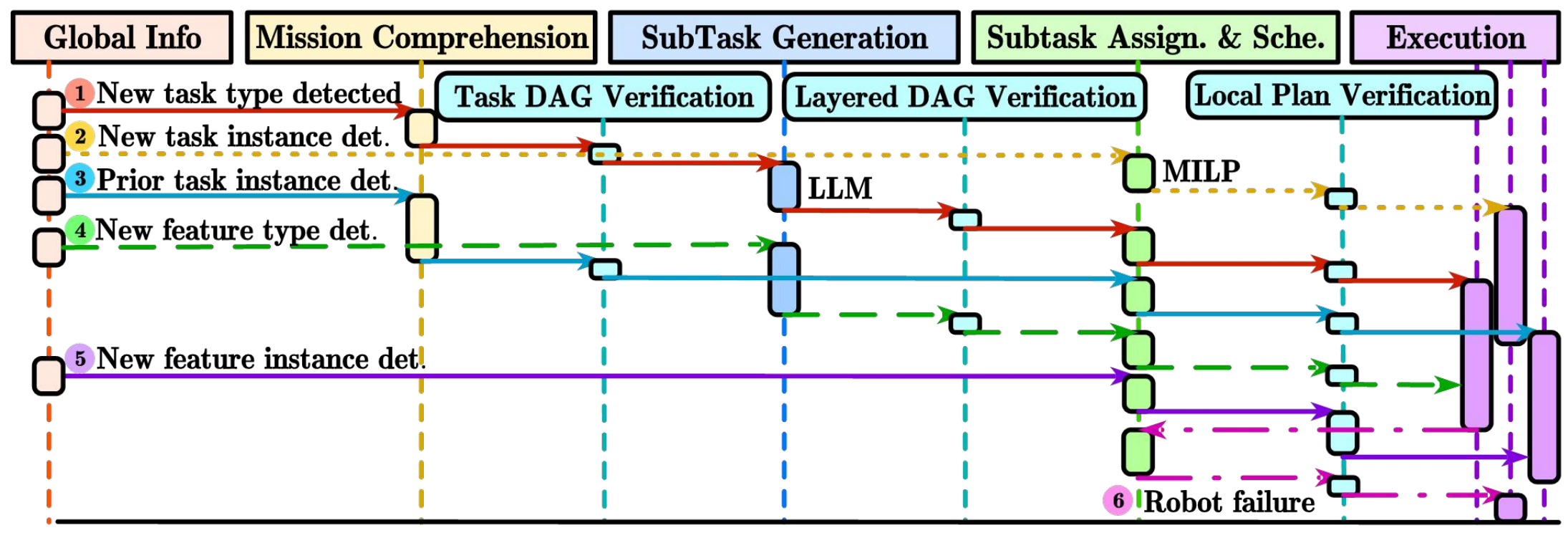}
    \vspace{-5mm}
    \captionsetup{font=footnotesize}
    \caption{\footnotesize Multi-level online adaptation to~$7$ types of events,
    and the human-in-the-loop verification after each module.}
    \label{fig:online-adaptation}
    \vspace{-7mm}
\end{figure}

The ability to adapt to different contingencies online is essential
for the robotic fleet
to operate in an open and unknown environment.
Particularly,
as shown in Fig.~\ref{fig:online-adaptation},
consider the following events that might appear
during the execution:
(I) \textbf{New instances of tasks detected}.
Whenever new instances of known tasks are detected (e.g., new victims detected),
the module of subtask assignment is re-triggered
to update the local plans of the robots.
(II) \textbf{New priority instances of tasks detected}
Whenever new priority instances of known tasks are detected (e.g., new serious victims detected),
the module of mission comprehension is re-triggered to update the task DAG.
Afterwards,
new nodes in the layered DAG are created by replicating
another node of the same type of task,
then the module of subtask assignment is re-triggered to update the local plans.
(III) \textbf{New types of tasks detected}.
Whenever new types of tasks are specified in a new mission,
the module of mission comprehension is re-triggered
to update the task DAG, which is then fed to
the following two modules to update the local plans
eventually.
(IV) \textbf{New types of features detected}.
Whenever new types of features are detected
(e.g., water besides fire extinguish),
it indicates that there are potentially new strategies
to decompose the task utilizing the new feature.
Thus, the module of subtask generation
is re-triggered to update the layered DAG,
which is then forwarded into the next module of subtask assignment
to potentially update the local plans.
(V) \textbf{New instances of features detected}.
Whenever new instances of known features are detected
(e.g., new reservoir detected),
it indicates that some subtasks might be accomplished earlier.
Thus, the map is updated with these features,
based on which the module of subtask assignment is re-triggered
to update the local plans.
(VI) \textbf{Subtask execution status updated}.
Whenever the robots accomplish a subtask,
this information is feedback to the module of subtask
assignment to monitor the progress of plan execution.
In case the execution of prolonged delay,
the task assignment is re-triggered to update the local plans.
(VII) \textbf{Robot failures}.
Whenever a robot fails,
it indicates that the current plan infeasible under the unchanged global information.
Thus, the module of task assignment is re-triggered to find the feasible local plans of the robots.

\begin{figure}[!t]
    \centering
    \includegraphics[width=1.0\linewidth, height=0.32\linewidth]{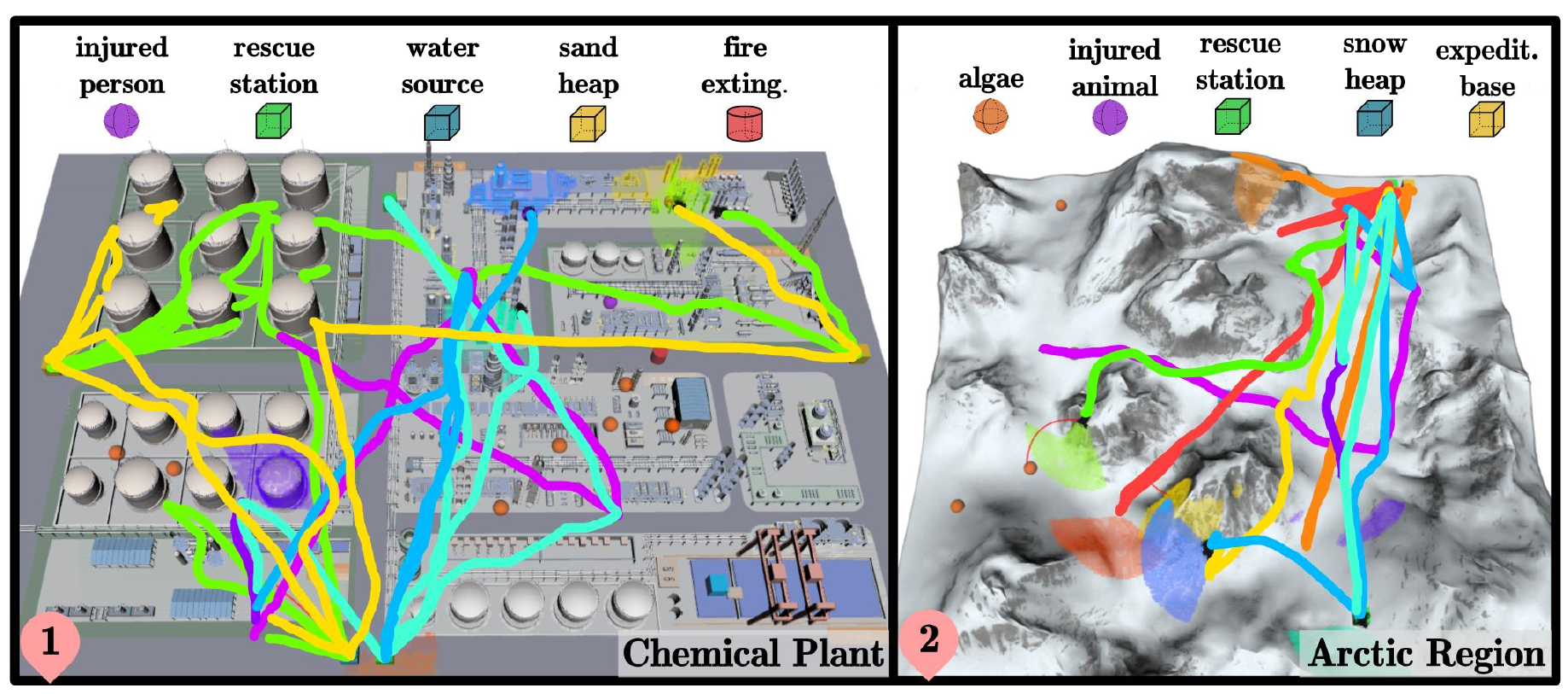}
    \vspace{-5mm}
    \captionsetup{font=footnotesize}
    \caption{\footnotesize Simulated scenarios:
    (I) Emergency at chemical plants with fire suppression and casualty rescue;
    (II) Preservation of arctic region with ecological remediation and wildlife rescue.
    }
    \label{fig:scenario}
    \vspace{-6mm}
\end{figure}

\begin{figure*}[!t]
    \centering
    \includegraphics[width=1.0\linewidth]{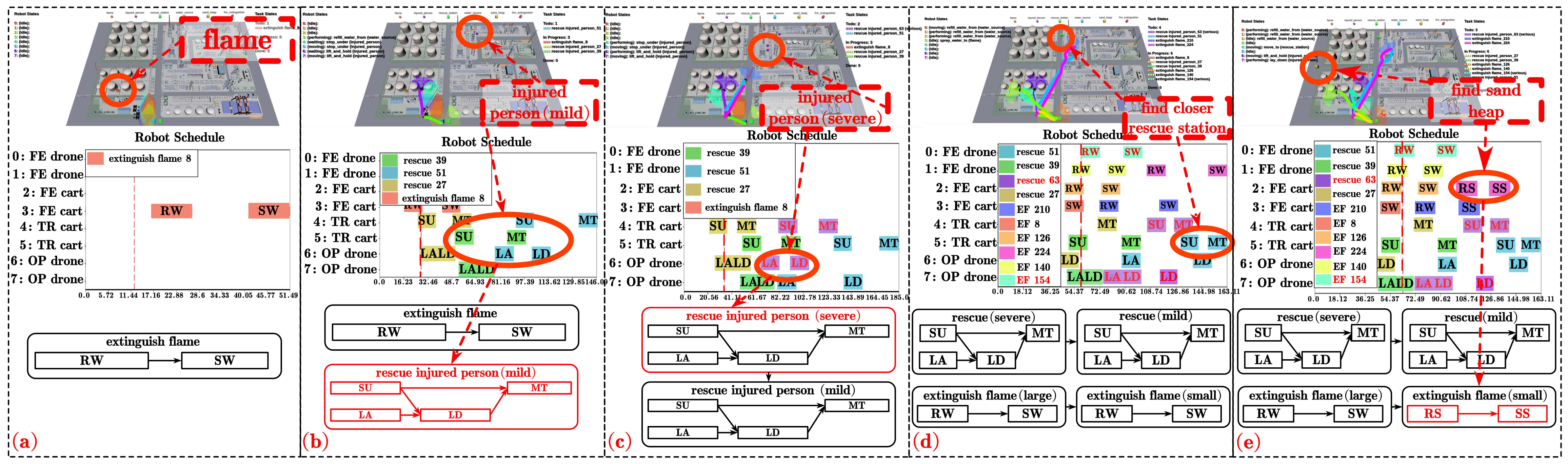}
    \vspace{-4mm}
    \captionsetup{font=footnotesize}
    \caption{\footnotesize Snapshots of results under Scenario-I:
      new tasks are detected (\textbf{(a)}-\textbf{(c)})
      new features  are detected (\textbf{(d)}-\textbf{(e)});
      new task DAG is added (\textbf{(b)}-\textbf{(c)});
      new task strategy is added (\textbf{(e)}).
      Note that the task assignment is updated accordingly after each update.
    }
    \label{fig:evol}
    \vspace{-5mm}
\end{figure*}

\subsubsection{Human-in-the-loop Verification} \label{subsubsec:verify}
To ensure correctness and improve transparency,
online verification and confirmation by the human operator with the basic knowledge of the mission
is enforced as follows:
the task DAG from the module of task comprehension
is verified regarding the partial ordering;
the layered DAG from the module of task decomposition
is checked regarding the strategies,
where human feedback can be provided as necessary;
the local plans from the module of subtask assignment
can also be verified regarding consistency with the constraints;
lastly, the execution status of the subtasks can be verified
regarding the actual outcome.
Theses measures have shown to be particularly effective
in case of complex dependencies among the tasks and subtasks,
where hallucination in LLMs can be apparent.
Thus, the intermediate and final outputs are all logically sound
and practically feasible,
enhancing robustness against errors in task planning and execution.

%% file: contents/experiment.tex
\section{Numerical Experiments} \label{sec:experiments}

To rigorously evaluate the proposed framework, 
numerical experiments are conducted across two large-scale scenarios. 
The implementation is built on~$\texttt{ROS}$ with~$\texttt{Python3}$, 
where the LLM module employs~$\texttt{DeepSeek-V3}$ for subtask generation 
(alternative models are compared in the sequel). 
All tests run on a workstation with an Intel i9-13900KF 24-Core CPU
@3.0GHZ with a RTX-4090 GPU. 
Simulation videos can be found in the supplementary material.

\input{contents/exp/setup.tex}
\input{contents/exp/result.tex}

\input{contents/exp/comp.tex}
\input{contents/exp/ablation.tex}
\input{contents/exp/general.tex}

%% file: contents/exp/setup.tex
\subsection{System Description}
\label{subsec:sys-desp}

Two large-scale scenarios are considered:
(I) \textbf{Industrial emergency} ($40\,\text{m} \times 60\,\text{m}$) contains dynamic hazards (expanding fires, immobile casualties)
and mission-critical resources (water reservoirs, extinguisher depots),
within which there are~$2$ manipulation UAVs, $2$ fire-suppression UAVs, $2$ casualty-transport UGVs,
and~$2$ multi-mode firefighting UGVs;
(II) \textbf{Arctic protection} ($45\,\text{m} \times 45\,\text{m}$) contains biohazard zones and wildlife,
maintained by~$4$ multi-functional UAVs, $2$ ice-surface UGVs, and $2$ heavy-payload UGVs.
Each robot is equipped with SLAM-based mapping, collision-free navigation, and task-specific actuators.
The proposed method is evaluated across tasks with diverse complexities:
\textbf{Mono-type} (single-robot task), \textbf{Dual-type} (cross-robot collaboration),
and \textbf{Compound-type} ($3$ robot coordination),
within each case from~$2$ to $8$ concurrent tasks are dynamically generated online.
Our framework is performed in a centralized manner,
so that the global information is available to all robots.
The ground truth (GT) strategy for each task is defined as the minimum sequences of subtasks
based on expert knowledge.
The system performance is quantified by deviations from the GT under spatio-temporal constraints.

%% file: contents/exp/result.tex
\subsection{Results under Scenario-I}
\label{subsec:result-eval}

The performance of the proposed framework is evaluated first in Secnario-I,
for the case of  compound tasks.
As shown in Fig.~\ref{fig:evol},
there are five key time instances that are worth pointing out.
At~$t=12$s in Fig.~\ref{fig:evol}a,
a new task of ``suppression fire'' is generated
with the strategy ``The firefighting UGVs first refill water (RW), then sprays water (SW) at the location'',
and triggers the module of subtask assignment,
which updates the local plans in~$1.0$s.
The detection of mild-injury victims at~$t=25$s in Fig.~\ref{fig:evol}b
invokes the full pipeline: mission comprehension in~$0.4$s,
subtask generation in~$60.1$s,
and schedule optimization in~$1.2$s,
where the new subtasks of ``coordinating parallel operations between the casualty-transport UGVs (SU)
and manipulation UAVs (LA) for victim access, 
followed by drone-assisted victim transfer (LD) to the vehicle and subsequent transport to the rescue station (MT)'',
are assigned to idle robots.
Then, at~$t=40$s in Fig.~\ref{fig:evol}c,
a severe-injury victim is detected
and the task DAG is reconstructed in~$0.5$s with priority constraints,
by leveraging existing strategy of rescue tasks.
The revised local plans show that this high-priority task is executed first
by Cart4 and Drone6,
while the other subtasks are re-assigned.
Moreover, a new rescue station is detected at~$t=50$s in Fig.~\ref{fig:evol}d
which only triggers the module of subtask assignment in~$3$s.
It results in a~$13.5\%$ reduction in task completion time,
as the victims are transferred to the nearest rescue station.
Lastly, novel resources like sand heaps are detected at~$t=60$s
in Fig.~\ref{fig:evol}e,
when the module of subtask generation triggers
the LLM-based reasoning and the strategy for~``suppression fire''
is updated to~``The firefighting UGVs first refill heap (RS), then sprays heap (SS) at the location''.
A total of~$30$ compound tasks and~$82$ subtasks are completed, fully satisfying all temporal-logical constraints. 
The average adaptation time across all events is \SI{66.21}{\second}, 
with the task DAG updated~$4$ times and the layered DAG updated~$5$ times on average.

%% file: contents/exp/comp.tex
\subsection{Comparisons}
\label{subsec:result-comp}

\begin{table*}[!t]
    \centering
    \renewcommand{\arraystretch}{1.2}
    \begin{tabularx}{\linewidth}{|p{3.5em}|p{6.5em}||D|E|C|C|C||C|E|C|C|C|C|}
    \hline
         & \multirow{2}{*}{\vspace{-1.1em}Methods} & \multicolumn{5}{c||}{Scenario-I} & \multicolumn{5}{c|}{Scenario-II} \\
    \cline{3-12}
         & ~ & SR$\uparrow$ & SPL$\uparrow$ & Time $\downarrow$ & Length$\downarrow$ & Tasks$\uparrow$ & SR$\uparrow$ & SPL$\uparrow$ & Time $\downarrow$ & Length$\downarrow$ & Tasks$\uparrow$ \\
    \hline \hline
        \multirow{6}{*}{\hspace{-0.5em}Mono}
            & CMRS & 1.00 & 0.16 & \underline{$\bm{22.01}$} & 12.33 & 30.27 & 1.00 & 0.22 & \underline{$\bm{24.78}$} & 9.33 & 38.74 \\
            & SMART-LLM & 1.00 & 0.20 & 44.71 & 10.33 & 33.00 & 1.00 & 0.25 & 64.42 & 8.67 & 35.58 \\
            & ROCO & 1.00 & 0.17 & 40.23 & 12.00 & 29.39 & 1.00 &
             0.17 & 48.98 & 13.33 & 26.16 \\
            & COHERENT & 1.00 & 0.32 & 37.07 & 6.33 & 72.96 & 1.00 & \underline{$\bm{0.44}$} & 40.00 & \underline{$\bm{4.67}$} & 86.96 \\
            & LiP-LLM & 1.00 & \underline{$\bm{0.65}$} & 102.37 & \underline{$\bm{3.50}$} & 62.55 & 1.00 & \underline{$\bm{0.44}$} & 89.13 & \underline{$\bm{4.67}$} & 60.93 \\
            & \textbf{Ours} & \underline{$\bm{1.00}$} & 0.63 & 61.98 & 3.67 & \underline{$\bm{166.60}$} & \underline{$\bm{1.00}$} & 0.40 & 78.53 & 5.50 & \underline{$\bm{114.49}$} \\
            \hline \hline
            \multirow{6}{*}{\hspace{-0.5em}Dual}
            & CMRS & 0.76 & 0.44 & \underline{$\bm{18.06}$} & 9.33 & 39.78 & 1.00 & 0.39 & \underline{$\bm{23.24}$} & 10.67 & 34.49 \\
            & SMART-LLM & 0.78 & 0.15 & 41.25 & 8.00 & 41.45 & 0.33 & 0.17 & 45.69 & 8.33 & 39.37 \\
            & ROCO & 0.67 & 0.22 & 45.93 & 11.67 & 29.62 & 0.33 & 0.11 & 57.00 & 11.33 & 29.38 \\
            & COHERENT & 0.67 & 0.33 & 67.92 & 8.00 & 51.84 & 1.00 & 0.56 & 87.64 & 7.33 & 49.35 \\
            & LiP-LLM & 1.00 & 0.50 & 64.17 & 8.00 & 52.86 & 0.33 & 0.17 & 54.95 & \underline{$\bm{6.00}$} & 66.69 \\
            & \textbf{Ours} & \underline{$\bm{1.00}$} & \underline{$\bm{1.00}$} & 70.45 & \underline{$\bm{4.00}$} & \underline{$\bm{153.68}$} & \underline{$\bm{1.00}$} & \underline{$\bm{0.56}$} & 75.31 & 7.33 & \underline{$\bm{87.65}$} \\
            \hline \hline
            \multirow{6}{*}{\hspace{-0.5em}Compound}
            & CMRS & 0.43 & 0.30 & \underline{$\bm{20.04}$} & 10.83 & 35.03 & 1.00 & 0.30 & \underline{$\bm{24.01}$} & 10.00 & 36.62 \\
            & SMART-LLM & 0.50 & 0.10 & 42.98 & 9.17 & 37.22 & 0.67 & 0.21 & 55.05 & 8.50 & 37.47 \\
            & ROCO & 0.83 & 0.19 & 43.08 & 11.83 & 29.51 & 0.67 & 0.14 & 52.99 & 12.33 & 27.77 \\
            & COHERENT & 0.83 & 0.33 & 52.49 & 7.17 & 62.40 & 1.00 & \underline{$\bm{0.50}$} & 63.82 & 6.00 & 68.15 \\
            & LiP-LLM & 1.00 & 0.57 & 83.27 & 5.75 & 57.71 & 0.67 & 0.31 & 72.04 & \underline{$\bm{5.33}$} & 63.81 \\
            & \textbf{Ours} & \underline{$\bm{1.00}$} & \underline{$\bm{0.82}$} & 66.21 & \underline{$\bm{3.83}$} & \underline{$\bm{160.14}$} & \underline{$\bm{1.00}$} & 0.48 & 76.92 & 6.42 & \underline{$\bm{101.07}$} \\
    \hline
    \end{tabularx}
    \captionsetup{font=footnotesize}
    \caption{\footnotesize {Quantitative comparison of different methods
        under varying tasks across two scenarios.
        Notes: SR (success rate (\%)),
        Time [s],
        Length: plan length,
        Tasks: number of completed tasks,
        SPL: step efficiency (generated vs. ground truth plan length).
        \underline{Best Result}.
        (↑: higher better, ↓: lower better).}}
    \label{tab:comp}
    \vspace{-5mm}
\end{table*}

To further validate the effectiveness of the proposed method,
comparative experiments are conducted against
five state-of-the-art approaches:
(I) \textbf{LiP-LLM}~\cite{obata2024lip} augmented with our module
of mission comprehension
for multi-robot allocation through linear programming and dependency graph integration;
(ii) \textbf{COHERENT}~\cite{liu2024coherent} modified to employ its
Proposal-Execution-Feedback-Adjustment loop exclusively
for LLM-based task assignment while retaining other core components;
(iii) \textbf{SMART-LLM}~\cite{kannan2024smart} enhanced with our mission comprehension framework for hierarchical planning
through LLM-guided coalition formation;
(iv) \textbf{CMRS}~\cite{huang2022language} implemented as an end-to-end baseline generating per-robot action plans;
and (v) \textbf{RoCo}~\cite{mandi2024roco} maintained in original configuration using multi-agent dialogues
for coordination and LLM-informed motion planning.
Five quantitative metrics are compared:
\textbf{Success Rate (SR)} as the percentage of trials producing valid, executable plans as verified by human experts,
\textbf{Plan Time (PT)} as computation time from environmental updates to plan generation,
\textbf{Planning Length} as the average number of discrete actions in generated plans,
\textbf{Tasks Completed} as the cumulative count of completed tasks,
and \textbf{SPL} computing step efficiency by comparing generated plan length to the ground truth~\cite{obata2024lip}.

The proposed method is evaluated in two scenarios,
each comprising three trials
with approximately~$200$ tasks and~$600$ subtasks within~$10000$s,
of which the results are summarized in Table~\ref{tab:comp}.
Our approach demonstrates consistently superior planning efficiency and task completion performance compared to baselines.
Notably achieving~$100\%$ success rates across all configurations,
our method completes significantly more tasks within identical time constraints (($166.60, 153.68, 160.14$) vs.~($114.49, 87.65, 101.07$)).
This is the most apparent in the case of compound task configurations
where all baselines fall short in both metrics of SR and SPL.
While baseline methods attain comparable SR in the mono-type tasks at~$100\%$,
our framework shows exceptional robustness in handling complex scenarios.
For the case of compound tasks, the highest SPL scores are achieved
by the proposed ($0.82$ vs.~$0.30$ vs.~$0.10$ vs.~$0.19$ vs.~$0.33$ vs.~$0.57$),
validating the effectiveness of the multi-stage subtask generation
via LLMs.
Although longer computation time is required than CMRS
(($61.98s, 70.45s, 66.21s$) vs. ($22.01s, 18.06s, 20.04s$)),
our method maintains the highest task efficiency,
owing to the model-based module of subtask assignment.
Despite of the performance degradation in Scenario-II compared
to Scenario-I (($114.49, 87.65, 101.07$)
vs. ($166.60, 153.68, 160.14$)),
our method retains the best performance in all metrics.

%% file: contents/exp/ablation.tex
\subsection{Ablation Study}
\label{subsec:result-ablation}

To better understand the contribution of different design choices in our framework,
ablation studies are conducted by analyzing the impact
of removing different stages in the module of subtask generation.

\begin{table}[t]
    \centering
    \renewcommand{\arraystretch}{1.1}
    \setlength{\tabcolsep}{1.5pt} 
    \begin{tabular}{l|cccc|cccc}
        \toprule
        \textbf{LLM} & \multicolumn{4}{c|}{Scenario-I} & \multicolumn{4}{c}{Scenario-II} \\
        \cline{2-9}
        \textbf{/Stages} & SR↑ & SPL↑  & Length↓ & Tasks↑ & SR↑ & SPL↑ & Length↓ & Tasks↑ \\
        \midrule
        DeepSeek-V3 & 1.00 & 0.82  & 3.83 & 160.14 & 1.00 & 0.48  & 6.39 & 101.18 \\
        GPT-4o & 1.00 & 0.79  & 4.25 & 170.81 & 1.00 & 0.91  & 3.50 & 206.35 \\
        qwen-2.5-max & 1.00 & 0.85  & 3.58 & 179.86 & 1.00 & 0.79  & 3.97 & 172.26 \\
        Grok-3 & 1.00 & 0.61  & 5.08 & 128.23 & 1.00 & 0.50  & 6.00 & 106.89 \\
        \midrule
        \midrule
        Ours w/o S.A & 1.00 & 0.61 & 5.33 & 110.44 & 1.00 & 0.46  & 6.33 & 98.76 \\
        Ours w/o M.P & 1.00 & 0.47 & 6.33 & 103.86 & 1.00 & 0.40  & 7.42 & 87.60 \\
        Ours w/o S.G & 1.00 & 0.72 & 4.33 & 124.43 & 1.00 & 0.47  & 6.50 & 101.18 \\
        \textbf{Ours (full)} & $\bm{1.00}$ & $\bm{0.82}$ & $\bm{3.83}$ & $\bm{160.14}$ & $\bm{1.00}$ & $\bm{0.48}$ & $\bm{6.39}$ & $\bm{106.29}$ \\
        \bottomrule
    \end{tabular}
    \captionsetup{font=footnotesize}
    \caption{\footnotesize Ablation studies in both scenarios:
      different LLMs (\textbf{upper}) and without certain stages
      in the module of task decomposition (\textbf{bottom}).
    }
    \label{tab:ablation}
    \vspace{-7mm}
\end{table}

\subsubsection{Different LLMs}
\label{subsub:aba-models}
Besides DeepSeek-V3,
other known LLMs are tested, such as GPT-4o, qwen-2.5-max, Grok-3.
As shown in Table~\ref{tab:ablation},
GPT-4o and Qwen-2.5-max achieve superior performance across both scenarios in terms of SPL, the quality of local plans,
and number of completed tasks.
Notably, our framework maintains high efficiency even for
models like Grok-3,
with~$100\%$ success rate and slightly degraded performance in SPL and completed tasks.
Moreover, it is worth noting that
the task efficiency exhibits variances across different LLMs.
In Scenario-I, Qwen-2.5-max achieves the peak SPL of~$0.85$
with the minimal plan length of~$3.58$,
whereas GPT-4o has the highest number of completed tasks at~$206.35$.
This observation suggests that the performances of different LLMs
are scenario-dependent.

\subsubsection{Different Stages within Module of Task Decomposition}
\label{subsub:aba-components}
To validate the importance of different stages within
the module of task decomposition,
ablation studies by sequentially removing Situation Analysis (S.A.),
Meta Policy (M.P.), and Subtask Generation Guide (S.G.).
As shown in Table~\ref{tab:ablation},
while all variants maintain perfect success rates (SR = 1.00),
their removal distinctly impacts the performance metrics.
The M.P. emerges as the most critical component, exhibiting the severest performance degradation: 42.7\% lower SPL (0.47 vs.~0.82), 65.3\%
longer average path length (6.33 vs.~3.83),
and 35.2\% fewer completed tasks (103.86 vs.~160.14) compared to the full implementation. S.A. demonstrates the secondary importance,
while S.G. shows the least impact.
This hierarchy aligns with our architectural priorities:
The M.P. drives the core planning throughout execution,
while S.A. ensures accurate scene interpretation,
and S.G. primarily optimizes LLM outputs.
These results confirm that the multi-stage reasoning capability (enabled by M.P.)
and the precise situation understanding (provided by S.A.)
constitute essential requirements for complex task decomposition.

%% file: contents/exp/general.tex


\subsubsection{Triggering of Different Modules during Online Adaptation}
\label{subsubsec:online}
As an important feature of the proposed module of online adaptation,
the modules of mission comprehension (MisComp), subtask generation (SubGen),
and scheduling (SubAll) are triggered at different events.
As shown in Table.~\ref{tab:verify},
$30$ tasks and~$10$ features are detected online for Scenario-I,
for which~$19\%$ triggers MisComp, $29\%$ for SubGen,
and $100\%$ for SubAll;
$20$ tasks and $8$ features are detected online for Scenario-II,
for which~$38\%$, $53\%$, and $100\%$ triggers MisComp, SubGen
and SubAll, respectively.
This reduces the LLM usage by~$81\%$ in Scenario-I and $62\%$ in Scenario-II,
compared to the baseline methods that query LLMs at each event.
Minimizing dependence on LLMs enhances cost-efficiency and ensures reliability,
for time-sensitive robotics applications in constrained environments.

\subsubsection{Importance of Human Verification}
\label{subsubsec:verify}
Furthermore,
the importance of online human-in-the-loop verification
is verified by removing verification after each module.
As shown in Table.~\ref{tab:verify},
in Scenario-I, the SR reaches~$100\%$ with verification,
compared to only~$65\%$ without.
Similarly in Scenario-II, the SR achieves~$100\%$ with verification,
while it drops to~$71\%$ without.
On average, only~$3$ interventions
are required during~$11$ tests for Scenario-I,
and $2$ of $7$ tests for Scenario-II,
to achieve~$100\%$ success rate.
Thus, these findings validate the importance of online human verification
and the seldom interventions actually required.

\begin{table}[!t]
    \centering
    \setlength{\tabcolsep}{4pt}
    \begin{tabular}{|c|c c c|c||c|c|c|}
        \hline
        \multicolumn{1}{|c|}{\multirow{2}{*}{Scenario}}                                    & \multicolumn{3}{c|}{Ours}                                                                                        & \multicolumn{1}{c||}{\multirow{2}{*}{Others}} & \multirow{2}{*}{} & \multirow{2}{*}{w/} & \multirow{2}{*}{w/o} \\ \cline{2-4}
                                    & \multicolumn{1}{c|}{MisComp}                & \multicolumn{1}{c|}{SubGen}                & SubAll              & \multicolumn{1}{c||}{}                        &                   &                     &                      \\ \hline
        \multirow{2}{*}{I} & \multicolumn{1}{c|}{\multirow{2}{*}{19\%}} & \multicolumn{1}{c|}{\multirow{2}{*}{29\%}} & \multirow{2}{*}{100\%} & \multirow{2}{*}{100\%}                       & SR                & 100\%               & 65\%                 \\ \cline{6-8}
                                    & \multicolumn{1}{c|}{}                      & \multicolumn{1}{c|}{}                      &                        &                                              & Ct.              & 3/11                 & ---                    \\ \hline
        \multirow{2}{*}{II} & \multicolumn{1}{c|}{\multirow{2}{*}{38\%}} & \multicolumn{1}{c|}{\multirow{2}{*}{53\%}} & \multirow{2}{*}{100\%} & \multirow{2}{*}{100\%}                       & SR                & 100\%               & 71\%                 \\ \cline{6-8}
                                    & \multicolumn{1}{c|}{}                      & \multicolumn{1}{c|}{}                      &                        &                                              & Ct.            & 2/7                & ---                    \\ \hline
    \end{tabular}
    \captionsetup{font=footnotesize}
    \caption{\footnotesize
    {\textbf{Left}: comparison of counts for different modules being triggered between ours with other baselines
    during the online adaptation of~$40$ events in Scenario-I
    and~$28$ events in Scenario-II;
    \textbf{Right}: success rate with and without the
    proposed human-in-the-loop verification,
    and the average number of feedback required.}
    }
    \label{tab:verify}
    \vspace{-7mm}
\end{table}

%% file: contents/conclusion.tex
\section{Conclusion and Future work} \label{sec:conclusion}

This work introduces DEXTER-LLM,
a novel framework enabling dynamic coordination of heterogeneous multi-robot systems in unknown environments.
Via combining LLM-based hierarchical reasoning with model-based optimization,
the framework achieves automated task decomposition,
constraint-aware scheduling, and online adaptation to emergent features,
while ensuring optimality and explainability.
Future work include the consideration of
inter-robot communication constraints,
and hardware experiments.